\documentclass[aps,pre,floatfix,showpacs,nofootinbib]{revtex4}
\usepackage{amssymb}
\usepackage[]{amsmath}
\usepackage{IEEEtrantools}
\usepackage{graphicx}
\usepackage{epstopdf}
\usepackage{booktabs}
\newcommand{\comment}[1]{}

\newcommand{\clz}{C_{\rm LZ}}

\newcommand{\BEQ}{\begin{equation}}
\newcommand{\EEQ}{\end{equation}}
\newcommand{\BEA}{\begin{eqnarray}}
\newcommand{\EEA}{\end{eqnarray}}

\newcommand{\ve}{\varepsilon}

\begin{document}
\draft
\title{Two halves of a meaningful text are statistically different}

\author{Weibing Deng$^{1)}$, R. Xie$^{1)}$, S. Deng$^{1)}$, and Armen E. Allahverdyan$^{2)}$ 
\footnote{wdeng@mail.ccnu.edu.cn, armen.allahverdyan@gmail.com}}

\address{ $^{1)}$Key Laboratory of Quark and Lepton Physics (MOE) 
and Institute of Particle Physics,  Central China Normal University, Wuhan 430079, China,\\
  $^{2)}$Yerevan Physics Institute, Alikhanian Brothers Street 2,
  Yerevan 375036, Armenia}


\begin{abstract}

Which statistical features distinguish a meaningful text (possibly
written in an unknown system) from a meaningless set of symbols?  Here
we answer this question by comparing features of the first half of a
text to its second half. This comparison can uncover hidden effects,
because the halves have the same values of many parameters (style, genre
{\it etc}). We found that the first half has more different words and
more rare words than the second half. Also, words in the first half are
distributed less homogeneously over the text in the sense of of the
difference between the frequency and the inverse spatial period. These
differences hold for the significant majority of several hundred
relatively short texts we studied. The statistical significance is
confirmed via the Wilcoxon test.  Differences disappear after random
permutation of words that destroys the linear structure of the text.
The differences reveal a temporal asymmetry in meaningful texts, which
is confirmed by showing that texts are much better compressible in their
natural way (i.e. along the narrative) than in the word-inverted form.
We conjecture that these results connect the semantic organization of a
text (defined by the flow of its narrative) to its statistical features.

\comment{
Several other features do not show a statistically
significant difference between the first and second halves:
repetitiveness of words (as quantified by Yule's constant), the length
of words, number of sentences, paragraphs {\it etc}. }

\comment{and they can be checked
without any understanding of the text.  They indicate on a higher
hierarchic level of text organization that so far went unnoticed in text
linguistics. }

\end{abstract}


%% 05.20.-y statistical mechanics
%% 05.65.+b self-organization in statistical mechanics
%% 05.10.Gg stochastic models in statistical mechanics and non-linear dynamics

%% 05.65.+b Self-organized systems +

%% 89.75.-k Complex systems (for complex chemical systems, see
%% 82.40.Qt; for biological complexity, see 87.18.-h)

%% 89.75.Da Systems obeying scaling laws +

%% 89.75.Fb Structures and organization in complex systems +

\maketitle

\section{Introduction}

\label{intro}

Texts are hierarchic constructs which consist of several autonomous
levels \cite{hutchins,valgina,hasan}: letters, words, phrases, clauses,
sentences, paragraphs. If a text is looked at statistically, i.e.
without understanding its meaning (e.g. because it is written in an
unknown system), how can it be efficiently distinguished from a
meaningless collection of words \cite{baa,orlov,arapov}? Several such
distinctions are well-known, e.g.  {\it (i)} meaningful texts have a
large number of words that appear in a text only few times, in
particular once (rare words or {\it hapax legomena}) \cite{baa}.  {\it
(ii)} Ranked frequencies of words obey the Zipf's law
\cite{estoup,condon,zipf}. {\it (iii)} Letters and words of a text
demonstrate long-range (power-law) correlations
\cite{lrc_schenkel,lrc_shnerb,lrc_ebeling,lrc_eckmann,lrc_manin,lrc_altmann}. 

However, these characteristics can be reproduced by a sufficiently
simple stochastic models putting in doubt their direct relation to the
meaningfulness of a text.  {\it (i)} Simple stochastic models can
recover quite precisely the detailed structure of the hapax legomena
\cite{pre}.  {\it (ii)} Zipf's law can be deduced from statistical
approaches
\cite{shrejder,li,simon,zane,kanter,hill,pre,liu,baek,vakarin,dover,latham,mandelbrot,mandel,manin}. The
first model of this type was a random text, where words are generated
through random combinations of letters, i.e. the most primitive
stochastic process \cite{mandelbrot,li}. Its drawbacks
\cite{howes,seb,cancho} (e.g. many words having the same frequency) are
avoided by more refined models \cite{simon,zane,kanter,hill,pre}.  
More generally, it was recently understood that Zipf's law is a statistical regularity
that emerges in samples which are informative about the underlying generative process \cite{cubero}. 
{\it (iii)} Physics and mathematics of stochastic processes offer a plethora
of models and approaches for generating long-range correlations
\cite{buck} \footnote{E.g. Ref.~\cite{lrc_schenkel} points out that
long-range correlations are found also in a dictionary, where the
meaning of text (as opposed to the meaning of words and phrases) is
absent. Ref.~\cite{lrc_manin} also argued against a direct relation between
long-range correlations and semantic structures.}. 

Here we contribute to resolving the above question by recalling that
meaningful texts evolve sequentially (linearly) from beginning to end.
This was taken as one of basic features of language \cite{sure},
which|together with other design features|allows to distinguish human
language from other communication systems \cite{hockett}. Thus we divide
texts into two halves, each one containing the same amount of words.
Thereby we neutralize confound variables that are involved in a complex
text-producing process (style, genre, subject, the author's motives and
vocabulary {\it etc}), because they are the same in both halves. Hence
by comparing the two halves with each other we hope to see statistical
regularities that are normally shielded by above variables.  Statistical
regularities hold for the majority of texts, such that it is highly
unlikely to get this majority for random reasons (as checked by the
$3\sigma$ rule).  The significance of results will be checked by
well-accepted statistical tests; for our purpose this is the W-test
(Wilcoxon's test) \cite{wilcoxon}. 

In two sets of several hundred texts we noted the following statistical
regularities.  {\it (1)} The first half has a larger number of different
words, i.e. a larger vocabulary. {\it (2)} It also has a larger number
of rare words, i.e.  words that appear once or twice. {\it (3)} The
first half is less compressible than the second half. The
compressibility was studied via several different standard approaches,
e.g. the Lempel-Ziv complexity and the zip algorithm. Lesser
compressibility relates to more information in the sense of Shannon
\cite{cover}.  {\it (4)} Common words of both halves tend to have a
larger overall frequency in the second half. These four features
significantly correlate with each other as quantified by Pearson's
correlation coefficient. {\it (5)} The words in the first half are
distribued less homogeneously, since they have a larger difference
between the frequency and (inverse) spatial period. 

One possible explanation of these result is that the first part of the
text normally contains the exposition (which sometimes can be up to 20
\% of the text), where the background information about events,
settings, and characters is introduced to readers. The first part also
plots the main conflict (open issue), whose denouement (solution) comes
in the second half \footnote{\label{foo1}Scientific texts contain
closely related aspects: introduction, critique of existing approaches,
statement of the problem, resolution of the problem, implications of the
resolution {\it etc}. The discussion on differences between the halves
applies also here.}. Hence {\it (1)-(4)} can be hypothetically explained
by the fact that the exposition|hence the first half|needs more
different words {\it (1)}, more rare words {\it (2)}, is more
informative (in the sense of Shannon) {\it (3)}, and introduces words that
are employed in the second half {\it (4, 5)} (i.e. the second half
processes information introduced in the first half).  We emphasize that
this explanation is hypothetical, its direct validity is yet to be
checked via more refined methods to be developed in future. 

{\it (6)} Many other features|in particular those related to
higher-order hierarchic structures of the text
\cite{hutchins,valgina,hasan}|do not show any significant difference
between the first and second half: number of sentences, paragraphs,
repetitiveness of words (as quantified by Yule's constant), number of
punctuation signs {\it etc}. Among such features we especially mention
the overall number of letters, since there is a weak statistical evidence
(the W-test is not always passed) that this quantity is still larger in
the first half. 

Checking these features does not require any understanding of the text,
i.e.  it is not required that the meaning of words is understood, or
even their writing system is known.  We show that (expectedly) neither
of them survives if the words of the text are randomly permuted, and only
after that the resulting ``text" is divided into two parts. Hence these
features are specific for meaningful texts and they can be employed for
distinguishing meaningful texts from a random collection of words. 

\comment{Moreover, they do no demand any text parsing, beyond
conservation of its natural (linear) order.  Not even the parsing into
words is necessary, since this can be done by detecting the space as the
most frequent symbol in the text. }

This paper is organized as follows. The next section reviews our data
collection method and recalls the W-test. Sections \ref{I}--\ref{spato}
present results from the above points {\it (1)-(5)}.
Section \ref{nego} reviews negative results from point {\it (6)}.  
The last section contains the outlook of this research
and relates it with existing literatures. Our main results
are briefly summarized in Table~\ref{tab0}. All other tables are given in
Appendix \ref{tablo}. 

%%%%%%%%%%%%%%%%%%%%%%%%%%%%%%%%%%

\begin{table}
\centering
\begin{tabular}{l|c|c} \hline\hline
&  First half & Second half \\
\hline
Number of different words; cf.~(\ref{udosh}) & + & -- \\
\hline
Number of rare words (absolute and relative); cf.~(\ref{cov}, \ref{boris}) & + & -- \\
\hline
Compressibility of the size; cf.~(\ref{compo}, \ref{sss}) & -- & + \\
\hline
The overall frequency of common words; cf.~(\ref{cc}) & -- & + \\
\hline
Difference between frequency and inverse spatial period; cf.~(\ref{ort})   & + & --  \\
\hline\hline
Number of letters; cf.~(\ref{guppi}) & + & -- \\
\hline\hline
Repetitiveness of words (Yule's constant); see (\ref{yule}) & $\emptyset $ & $\emptyset $ \\
\hline
Number of punctuation signs & $\emptyset $ & $\emptyset $ \\
\hline
Average length of words & $\emptyset $ & $\emptyset $ \\
\hline
Number of sentences & $\emptyset $ & $\emptyset $ \\
\hline
Average length of sentence & $\emptyset $ & $\emptyset $ \\
\hline
Number of paragraphs & $\emptyset $ & $\emptyset $ \\
\hline
Size in bytes  & $\emptyset $ & $\emptyset $ \\
\hline
%Number of functional words  & $\emptyset $ & $\emptyset $ \\
\hline
\end{tabular}
\caption{\label{tab0}Qualitative comparison of
various features of texts between first and second halves: + (--) means
that the feature is larger (smaller) in the corresponding half.
$\emptyset $ means that the sought difference does not show up.  
Features are divided into two
groups by double-lines. The first four features correlate with each other.
The number of letters is separated, 
since there is a weak evidence toward its validity (the 
values of test quantities are close to their threshold values).  }
\end{table}

%%%%%%%%%%%%%%%%%%%%%%%%%%%%%%%%%%%%

\section{Data collection and testing}

\subsection{Studied texts}

We selected English texts with a single narrative that are written on relatively few tightly
connected subjects, and are sufficiently short for not containing ``texts
inside of texts'' \footnote{To ilustrate this point, consider ``War and Peace'' by
Leo Tolstoy. This big novel is written in two different languages
(Russian and French), and contains a big amount of heroes and
circumstances. It does have several parallel narratives, and describes
situations in the course of twenty years.  Clearly, this is a text of
texts, and it would not be meaningful to focus on dividing it over two
halves. But we can divide over two halves one of its (long) chapters. We
have not done this so far.  }.
We divided studied texts into two halves (each half 
contains equal number of words) \footnote{We got a
preliminary evidence that, as expected, dividing texts into more than two
parts will obscure the text difference effects shown in Table I. Note that
Ref.~\cite{zano} studies text division into several parts, but that was
done for a different purposes. For long texts containing
$155000-220000$ words, Ref.~\cite{zano} noted that such texts can be divided into several
sub-texts of the size of $1000-3000$ words. The criterion of separation
is qualitatively close to the above concept of ``books inside
of a book", because it looked at the spatial clustering of key-words.
I.e. a group of key-words appearing mostly in one part of the text and
not in others, will effectively define a sub-text.} along the flow of the narrative, i.e. from
the beginning to end. Several aspects of texts are left unchanged: each half
is sufficiently large for statistics to apply, they
have the same overall number of words, the same author, genre {\it etc}.
The halves are semantically different, since the first half can be understood
without the second half, but the second half generally cannot be
understood alone. Also, the structure of narrative is different: the
first half normally contains the exposition, where actors, situations
and conflicts are set and defined, while the second half normally
contains the denouement; cf.~Footnote~\ref{foo1}. 

We have chosen to work with two datasets \cite{online}. The first dataset was taken
from the Gutenberg project at \cite{gutenberg}. It
consists of 156 fiction novels; for each novel the overall number of
words is in the range $ [10000, 50000]$, which is sufficiently large for
statistics to apply, but sufficiently short to ensure that they do not
amount to ``books inside of a book''.  This range of the overall number
of words $ [10000, 50000]$ is motivated from our own experience as
readers. 

The texts within the first dataset are thematically close, since they
are all fiction novels. The second dataset consists of 350 thematically
diverse texts taken from various online sources and collected at
\cite{ting}.  When collecting those texts we tried to ensure that they
do not contain texts that are meaningless to divide into halves, i.e.
that they do not contain effectively independent narratives. Hence we
did not include in this dataset biographies, poems, collections of short
stories or essays (in particular, folk stories), lectures, proceedings,
letters. 

\subsection{Testing the difference between the halves}

After processing, the typical form of our data are pairs of numbers for
each text: $\{x_{1,i}, x_{2,i}\}_{i=1}^M$, where $x_{1,i}$ and $x_{2,i}$
are certain features of (resp.) the first and second half of a text $i$,
with $M$ being the overall number of texts in the dataset. E.g. $x_{1,i}$ and $x_{2,i}$
are the number of different words for (resp.) first and second halves of a text; see below.

To inquire on
whether this data indicate on a difference between two halves, we
formulate two natural hypotheses: ${\cal H}_1$ (${\cal H}_0$) means that
the difference $x_{1,i}-x_{2,i}$ does (does not) follow a symmetric
distrbution around the zero. Now some understanding on excluding ${\cal H}_0$ can be gained
by looking at the percentage of cases, where $x_{1,i}<x_{2,i}$. This amounts to calculating
\BEA
\label{gnu1}
\sum_{k=1}^M{\rm sgn}[x_{2,i}-x_{i,i}].
\EEA
But looking only at (\ref{gnu1}) is incomplete, since it does not take into account the magnitudes 
$|x_{1,i}-x_{2,i}|$. A relatively complete answer to the above comparison between 
${\cal H}_1$ and ${\cal H}_0$ is provided by the W-test (Wilcoxon's test) 
\cite{wilcoxon}, which looks at [cf.~(\ref{gnu1})]
\BEA
\label{gnu2}
W\equiv\sum_{k=1}^M R_i\,{\rm sgn}[x_{2,i}-x_{i,i}], \qquad (R_1,...,R_M)
={\rm ordered}(\, \{|x_{1,i}- x_{2,i}|\}_{i=1}^M\,),
\EEA
where $\{|x_{1,i}- x_{2,i}|\}_{i=1}^M$ is ordered in increasing way, and asigned ranks $R_i$
that enter into $W$ \cite{wilcoxon}. Note that $W$ does not depend on absolute magnitudes, i.e. $W$ is invariant upon
multiplying $x_{1,i}$ and $x_{2,i}$ by the same positive number. Now if ${\cal H}_0$ holds, then for $M\gg 1$ 
(effectively $M>30$ suffices, which always holds in our cases) the law of large numbers works and
$W$ is a Gaussian random variable, since it is a weigted sum of a large number of uncorrelated random variables. 
Its average is zero, since ${\rm sgn}[x_{2,i}-x_{i,i}]$ assume values $\pm 1$ with equal probability (once 
${\cal H}_0$ is assumed to hold). Its dispersion is calculated directly from (\ref{gnu2}) \cite{wilcoxon}:
\BEA
\label{sigma}
\sigma_W^2(M)=\langle W^2\rangle=\sum_{k=1}^M k^2=\frac{M(M+1)(2M+1)}{6}.
\EEA
Hence ${\cal H}_0$ can be excluded via the $3\sigma$ rule, if 
\BEA
\label{3sigma}
|W|>3\sigma_W (M).
\EEA
We accept the $3\sigma$ rule (\ref{3sigma}) as the minimal threshold for claiming the statistical
significance of our results. However, we emphasize that the absolute majority of
our results hold the much stronger $5\sigma$ rule; see tables in Appendix \ref{tablo}. 

\section{Words: different, rare, common}
\label{I}

\subsection{Different words (vocabulary)}

The basic hierarchic level of text is that of words. Neglecting phenomena
of synonymy and homonymy (which are rare in English, but not at all rare
e.g. in Chinese \cite{epjb}), we can say that every word has several
closely related meanings (polysemy). Neglecting also the difference
between polysemic meanings, the number of independent meanings in a text
can be estimated via the number of different words $n$. Tables
\ref{n_156} and \ref{n_350} show that the first half of a meaningful
text has statistically more different words than the second half:
\BEA
\label{udosh}
n_1>n_2
\EEA
As expected, this result disappears after random shuffling (random
permutation of words) of texts that destroys its linear structure; see
Tables \ref{shuffle_156} and \ref{shuffle_350}. 

\subsection{Rare words (hapax legomena)}

In any meaningful text, a sizable number of words appear only very
few times ({\it hapax legomena}).  These rare words amount to a finite
fraction of $n$ (i.e. the number of different words).  The existence and
the (large) number of rare events is not peculiar for texts, since there
are statistical distributions that can generate samples with a large
number of rare events \cite{baa,pre}. One reason
why many rare words should appear in a meaningful text is that a typical
sentence contains functional words (which come from a small pool), but
it also has to contain some rare words, which then necessarily have to
come from a large pool \cite{latham} \footnote{\label{lato}E.g. this sentence
contains rare words {\it typical} and {\it pool} that in the present text
are met only 3 and 2 times, respectively.  It also contains frequent
words {\it words}, {\it since}, {\it large}.}.  

\comment{\BEA
\label{kusho}
\sum_{m=1}^k V_m^{[1]} \geq \sum_{m=1}^k V_m^{[2]}, \qquad k=1,...,5,
\EEA
where $V_m^{[1]}$ ($V_m^{[2]}$) is the number of words that appear $m$ times in the first (second) half.
For the halves (\ref{kush}) is written as 
\BEA
\label{kusho1}
&&{\sum}_{m=1}^{f^{[\ell]}_1 N/2}\, V^{[\ell]}_m=n_\ell, \\
&&{\sum}_{m=1}^{f^{[\ell ]}_1N/2} \, mV^{[\ell]}_m=N/2,
\qquad \ell=1,2,
\label{kusho2}
\EEA
where $n_1$ ($n_2$) is the number of different words in the first (second) half, and $f^{[1]}_1$ ($f^{[2]}_1$)
is the frequency of the most frequent word in the first (second) half. 
}

Let $V_m^{[1]}$ ($V_m^{[2]}$) is the number of words that appear $m$ times in the first (second) half.
For defining rare words we focused on 
\BEA
\label{hh}
h_\ell(\kappa) \equiv \sum_{m=1}^\kappa V_m^{[\ell]}\qquad \ell=1,2,\qquad \kappa=1,...,5,
\EEA
i.e. on words that appear up to $\kappa$ times. We choose to work with different $\kappa$'s to ensure that
our results are robust with respect to varying the definition of ``rare''.
For both datasets we observed that in the majority of cases the number
of rare words in the first half is larger than the number of rare words 
in the second half [see Tables~\ref{rare_156} and \ref{rare_350}]:
\BEA
\label{cov}
h_1(\kappa)>h_2(\kappa),\qquad \kappa=1,...,5.
\EEA
We confirmed via the $5\sigma$ of the W-test that the probability to get (\ref{cov}) due to random reasons is
negligible. 

Eq.~(\ref{cov}) suggests that the first half uses more rare words, but
such a conclusion is incomplete, since the two halves have different
numbers of distinct words. Denote them as $n_1$ and $n_2$, for the first
and second half respectively; cf.~(\ref{udosh}). 
Note that
\BEA
\label{kusho1}
{\sum}_{m=1}^{f^{[\ell]}_1 N/2}\, V^{[\ell]}_m=n_\ell, 
\qquad \ell=1,2,
\label{kusho2}
\EEA
where $f^{[1]}_1$ ($f^{[2]}_1$)
is the frequency of the most frequent word in the first (second) half. 
Hence in addition to (\ref{cov}) it is necessary to consider normalized
quantities, i.e.
\BEA
\label{boris} 
h_1(\kappa)/n_1>h_2(\kappa)/n_2, \qquad \kappa=1,...,5.
\EEA
Relation (\ref{boris}) does hold statistically; see
Table~\ref{rare_156} for the first dataset and Table~\ref{rare_350} for the
second dataset.  

Hence the first half has more rare words both in absolute and relative
terms; see (\ref{udosh}, \ref{boris}). These differences between the
halves disappear after random shuffling of texts; see Tables
\ref{shuffle_156} and \ref{shuffle_350}. 

Note that yet another possibility to define rare words comes from relations
\BEA
\label{kus}
{\sum}_{m=1}^{f^{[\ell ]}_1N/2} \, mV^{[\ell]}_m=N/2,
\qquad \ell=1,2.
\EEA
Eq.~(\ref{kus}) invites to compare the normalized quantities
$\frac{2}{N}\sum_{m=1}^\kappa m V_m^{[1]}$ with $\frac{2}{N}\sum_{m=1}^\kappa m V_m^{[2]}$ for $\kappa=1,...,5$.
We carried out this comparison and the results (for percentages and $W$-values) are very similar to (\ref{cov}), i.e.
we obtain
\BEA
\label{kookoo1}
\sum_{m=1}^\kappa m V_m^{[1]} > \sum_{m=1}^\kappa m V_m^{[2]}, \qquad \kappa=1,...,5,
\EEA
in the same statistical sense as (\ref{cov}).

\subsection{Common words}

Both halves of a text have certain common words, e.g. non-common words of the 
second half are those that are not met in the first half. Let the number of 
common words in a given text is denoted by $C$. Our first result is that for each half
the common words are less numerous than non-common ones:
\BEA
\label{urartu}
C<n_1/2,\qquad C<n_2/2,
\EEA
where $n_1$ ($n_2$) is the number of different words in the first (second) half.
Relations (\ref{urartu}) hold statistically; see Table~\ref{urartu_t}. Though common words
are in minority they are more frequent, i.e. the overall frequency $c_1$ ($c_2$) 
of common words in each half holds
\BEA
\label{van}
c_1>1/2,\qquad c_2>1/2.
\EEA
Relations (\ref{van}) hold without exclusions for all text we studied.
Inequalities (\ref{urartu}, \ref{van}) are well expected, since common
words include functional words, which are frequent, but not numerous
\cite{pre}. 

Our next finding does indicate on a difference between
two halves, and is therefore less expected: the overall frequency of common
words is larger in the second half [see Tables \ref{n_156} and
\ref{n_350}]:
\BEA
\label{cc}
c_1<c_2.
\EEA
Eq.~(\ref{cc}) indicates in which specific sense the second half employs more 
words from the first half, than {\it vice versa}. 

\section{Compressibility }

\subsection{Lempel-Ziv (LZ) complexity and compressibility }

Here we discuss to which extent the two halves differ by their
compressibility, and whether the word-inverted text has has a
compressibility compared to the original text. Compressibility|i.e. the
ability of size decreasing under lossless compression|is an interesting
indicator of textual features, e.g. it is known that a random
permutation of words in a text decreases its compressibility
\cite{lande}. The idea that texts should have a larger compressibility
than a random data also appeared in the form of Hilberg's conjecture
\cite{debowski1,debowski2}. Compressibility was recently employed for
detecting ordered structures in data \cite{gurzadyan}. 

Now the compressibility can be defined via one of
standard lossless compression methods e.g. the zip. But algorithms 
for such methods are not freely available. Hence we employ the
Lempel-Ziv (LZ) complexity which estimates compressibility 
and which is basic for other compression methods (zip, gzip) 
\cite{cover}. The LZ-complexity is widely
applied in data science; see e.g. \cite{lzlz} for a recent review.

Let us illustrate how the LZ-complexity is calculated
\cite{cover,lzlz} for a bit-string $01001011$. This string is separated
into fragments by commas starting from the left.
Each fragment is defined to be the shortest substring that did not appear
before: $01001011\to 0,1,00,10,11$. Now each fragment whose length is
larger than one bit consists of the last bit and the prefix, i.e.  the
part that already appeared somewhere later. (The last
fragment need not have the last symbol, and can simply consist of a prefix.) In the second step each
fragment is coded via the number of the fragment, where its prefix first
appeared (first symbol) and its last bit (second symbol). Thus
\BEA
01001011\to 0,1,00,10,11 \to (0;0) (0;1) (1;0) (2;0) (2;1).
\EEA
The LZ-complexity $\clz$ of the string is defined as the overall number
of fragments.  Now $\clz$ determines the bit-length of the coded string,
since we need $\clz$ bits for last symbols for each fragment plus (at
best) $\clz\log_2 \clz$ bits for representing prefixes \cite{cover}.
Obviously, $\clz$ can be significantly smaller than the initial string length
only for sufficiently long strings. The LZ-complexity defines a
universal (since no prior information about the string is to be available) and
asymptotically optimal lossless compression, since its compression ratio for a
{\it stationary random process} tends to the entropy rate of the process
\cite{cover}. 

$\clz$ reads for two simplest $N$-bit strings:
\BEA
\label{jiu1}
&&\clz(000000...0000)=\frac{1}{2}(\sqrt{8N+1}-1)\simeq \sqrt{2N},\quad {\rm for}\quad  N\gg 1,\\
&&\clz(010101...0101)=\sqrt{1+4N}-1\simeq 2\sqrt{N},
\label{jiu2}
\EEA
where (\ref{jiu1}) [(\ref{jiu2})] is derived from noting that the
lengths of successive fragments are $1,2,3,....$
[$1,1,2,3,2,3,4,5,4,5,6,7,6,7....$]. Eqs.~(\ref{jiu1}, \ref{jiu2}) are
to be contrasted with the length of optimal compressions for the
corresponding strings: $(0;N)$ and $(01;N/2)$. These lengths reduce to
representing $N$ via bits and amount to $1+\log_2 N$ for both cases. 
Though $\clz$ is far from the optimal case for ordered strings 
(\ref{jiu1}, \ref{jiu2}), it is interesting to see that the string in
(\ref{jiu2}) is still more LZ-complex, as expected intuitively. Thus,
more regularity leads to a better compression, a general idea of all lossless
compression methods. 

We transform each text into a bit-string 
\footnote{This is done in the most standard way:
letters and punctuation signs are replaced by their ASCII numbers
written in the binary representation.}, and define the relative compressibility as
\BEA
\label{compo}
s =({S_{\rm in}-\clz})/{S_{\rm in}},
\EEA
where $S_{\rm in}$ is the original size of a given text in bytes, and $\clz$ is the LZ-complexity. 
The normalization of $\clz$ by the original size of the string is standard \cite{cover}.

\subsection{Permutation and inversion}
\label{invo}

Note that if the string in (\ref{jiu2}) is randomly permuted
sufficiently many times, it will turn into a random string (with equal
number of $0$'s and $1$'s). The $\clz$ of such a random string will be
${\cal O}(N)$ \cite{cover}, i.e.  much larger than ${\cal O}(\sqrt{N})$
in (\ref{jiu2}). Results of \cite{lande} are easy to understand in this
context: random permutations will increase the LZ-complexity of any
given text. To confirm this relation in our databases we generated 10
random permutations of words in each text, after each permutation we calculated the difference of
compressibilities (\ref{compo}) between the original text and permuted
one, and then averaged the difference over 10 permutations. The
difference was positive for all our texts without exclusion.

Here is however a result that is much less straightforward. 
Once the present work focuses on the linear structure of a text, 
it is natural to ask the following
question: is there any compressibility difference between a given text
${\rm T}$ and its inverted version ${\rm T}_{\rm inverted}$? Here the
last word of ${\rm T}$ becomes the first word of ${\rm T}_{\rm
inverted}$, the penultimate word of ${\rm T}$ becomes the second word of
${\rm T}_{\rm inverted}$, ..., and the first word of ${\rm T}$ becomes
the last word of ${\rm T}_{\rm inverted}$. Next, ${\rm T}$ and and ${\rm
T}_{\rm inverted}$ are (separately) turned into bit-strings and their
compressibility is calculated via (\ref{compo}) \footnote{In this context we 
stress that the LZ-complexity is generally not symmetric with respect to
inverting the string. E.g. $\clz(0,01,010)=3$, but $\clz(0,1,01,00)=4$.}. 

It appears that the original text ${\rm T}$ is 
(statistically) more compressible in terms of (\ref{compo}) than the 
inverted text ${\rm T}_{\rm inverted}$ [cf.~Table \ref{inversion}]:
\BEA
\label{gel}
s>s_{\rm inverted}.
\EEA
The percentage of (\ref{gel}) is remarkably high: it holds for $>97\%$
cases in both of our datasets; see Table \ref{inversion}. 
Eq.~(\ref{gel}) is confirmed via the zip 
compression method; see Table \ref{inversion}. 

Recall that previous applications of the LZ-complexity in texts
\cite{lande,debowski1,debowski2} assumed that the LZ-complexity captures
correlations between different text symbols (letters, words {\it etc}).
Relation (\ref{gel}) can be explained by noting that $\clz$ focuses on
short range correlations of letters, which may be lost after inversion
of words. To illustrate this point, let $\ell_1\ell_2\ell_3 $ and
$\kappa_1\kappa_2\kappa_3 $ are two consecutive 3-letter words. Their
order in ${\rm T}$ [in ${\rm T}_{\rm inverted}$] is
$\ell_1\ell_2\ell_3\,\,\kappa_1\kappa_2\kappa_3 $
[$\kappa_1\kappa_2\kappa_3\,\,\ell_1\ell_2\ell_3$]. Now if there are
correlations between $\ell_3$ and $\kappa_1$ in ${\rm T}$, and such
correlations are accounted for in $\clz$, then in ${\rm T}_{\rm
inverted}$ such correlations will have a longer range, and will not be
seen in $\clz$ \footnote{\label{syntagma}Work is in progress to
understand whether such directional correlations are related to
syntagmatic correlations between words of the text well-known 
in linguistics \cite{sure,sahlgren}. Qualitatively, these are correlations
along the text determined by co-occurence of words or letters
\cite{sure,sahlgren}. They are conceptually different from paradigmatic
relations between the words, where two words having such a relation tend to
appear in the same context, i.e. in the same surrounding of words.}. 

Note that instead of inverting texts at the level of words we also
inverted them at the level of letters: put the last letter as the first
one {\it etc}. We saw that out of this letter inversion the
compressibility does change, i.e. there is more into the LZ-complexity
than just short-range correlations of letters. However, no clear
indications emerged on the analogue of (\ref{gel}) or on its inverse.
The results differ from one dataset to another and from one compression
method to another.

\comment{
Recall how the lossless compression methods (including zip)
work \cite{cover}: they look for various repeating patterns (within some
local window of consecutive symbols) and code more frequent patterns by
shorter codewords. Hence sequences that contain more repeating patters
are more compressible. }

\subsection{Compressibility of two halves}

Table \ref{zip} shows that the relative compressibility of the
first half is statistically smaller, i.e. it is compressed less than the second half:
\BEA
\label{sss}
s_1<s_2.
\EEA
Inequality (\ref{sss}) holds in 60\%-70\% of cases with at least $4\sigma$ significant $W$-factor.
The results are fully corroborated when using in (\ref{sss}) the zip compression method 
instead of the LZ-complexity. 

\comment{We emphasize that all the results below were confirmed via the
standard zip method.} 

Recall that according to information theory, more
compressible sequences convey less information (in the sense of Shannon)
\cite{cover}.  Then one can interpret (\ref{sss}) in the context of
(\ref{udosh}, \ref{cov}, \ref{boris}): the first half has more
words|hence it conveys more information|and more rare words, which
altogether combine into a more informative and hence less compressible
structure. Indeed Tables \ref{corr_156} and \ref{corr_350} show that the
validity of (\ref{sss}) does correlate (in the sense of the Pearson
correlation coefficient recalled in Appendix \ref{pearson}) with the
validity of (\ref{udosh}, \ref{cov}, \ref{boris}).

\section{Differences between the word frequency and the spatial period }
\label{spato}

\subsection{Definitions}

Let us now turn to features that reflect the distribution of words along
the text. Studying this spatial distribution of words is traditional for
quantitative linguistics \cite{zipf,yngve}. More recently,
Refs.~\cite{ortuno,pury,carpena,zano} investigated the spatial
distribution of key-words versus functional words. The conclusion
reached is that key-words are distributed less homogeneously, i.e. they cluster into certain parts of the text
\cite{ortuno,pury,carpena,zano}. 

For a given text we extract the frequencies of different words
\footnote{\label{janmuller}There is no universal definition of word
\cite{muller}; e.g. there is a natural uncertainty on whether to count
plurals and singulars as different words. Different definitions of word
can produce numerically different results \cite{muller}. We mostly work
with methods that assume singular and plural to be different words. But
we also checked that our qualitative conclusions do hold as well when
singular and plural are taken to be the same word.} ($n$ is the number of different words):
\BEA
\label{w1} \{f(w_r)\}_{r=1}^{n}, \qquad
{\sum}_{r=1}^{n} f(w_r) =1. \EEA

Let $w_{[1]},...,w_{[\ell]}$ denote all occurrences of a word
$w$ along the text. Let $\zeta_{\,i\,j}$ denotes the number of words
(different from $w$) between $w_{[i]}$ and $w_{[j]}$. I.e. 
$\zeta_{\,i\,j}+1\geq 1$ is the number of space symbols between 
$w_{[i]}$ and $w_{[j]}$. Define the
average period $t(w)$ of this word $w$ via
\begin{equation}
\label{durnovo}
t(w)=\frac{1}{\ell-1}{\sum}_{k=1}^{\ell-1} \,(\zeta_{\,k\,\,k+1}+1).
\end{equation}
The averaging is conceptually meaningful only for sufficiently frequent
words, though formally (\ref{durnovo}) is always well-defined.  Note
that $(\ell-1)t(w)$ equals to $1$ plus the number of words that differ
from $w$ and occur between $w_{[1]}$ and $w_{[\ell]}$. Hence $t(w)$ will
stay intact under redistributing $w_{[2]},...,w_{[\ell-1]}$ for fixed
$w_{[1]}$ and $w_{[\ell]}$. 

Now define the inverse spatial period:
\begin{equation}
\label{murnovo}
g(w)\equiv 1/t(w).
\end{equation}
If a word $w$ is distributed homogeneously, then $g(w)$ is expressed via
the ordinary frequency $f(w)$. If in addition, this is a sufficiently
frequent word, then $g(w)\approx f(w)=\ell/N$, where we assume that
$N\gg 1$ and $\ell\gg 1$. Hence the difference $|g(w)-f(w)|$ 
(for sufficiently frequent words) can tell how
the distribution of $w$ deviates from the homogeneous one. 

One can also directly define the average characteristic frequency
for the word $w$:
\begin{equation}
\label{vernovo}
\bar{\omega}(w)=\frac{1}{\ell-1}{\sum}_{l=1}^{\ell-1} \frac{1}{\zeta_{\,l\,\,l+1}+1}.
\end{equation}
One feature of (\ref{vernovo})|which is absent in (\ref{murnovo})|is that
(\ref{vernovo}) is not susceptible to outliers, e.g. if $\zeta_{12}$ is much larger
than other $\zeta_{i\,j}$'s, then $\zeta_{12}$ does not contribute much into (\ref{vernovo}).
Here we shall focus on $t(w)$, since (\ref{vernovo}) did not so far demonstrate any
interesting behavior \footnote{In particular, $g(w)$ in contrast to $\bar{\omega}(w)$ demonstrates
an analogue of the Zipf's law. Elsewhere we shall discuss this point in detail. }. 

\comment{
Eq.~(\ref{vernovo}) has the expected behavior for the
homogeneous distribution of $w$ within the text. For such distribution
all $\zeta_{i\,j}$ are equal: $\zeta_{i\,j}=\zeta$, where $\zeta$ is defined
from placing the word $w$ ($Nf(w)$ times and with equal intervals) among $N$
words. Hence $Nf(w)+(Nf(w)+1)\zeta=N$ and
$\bar{\omega}(w)=\frac{1}{\zeta+1}=\frac{\frac{1}{N}+f(w)}{\frac{1}{N}+1}$. 
Whenever $f(w)\gg \frac{1}{N}$ (and naturally $1\gg \frac{1}{N}$) we get
$\bar{\omega}(w)=f(w)$, i.e. the space frequency coincides with the ordinary one. 
It is seen that the largest value $\bar{\omega}=1$ of (\ref{vernovo}) is
achieved for $\zeta_{\,i\,\,i+1}=0$ when all occurences the word $w$
come after each other without any other word in between. The smallest
value of $\bar{\omega}=\frac{1}{N-1}$ is achieved for
$\zeta_{\,1\,\,2}=N-2$ with just two occurences of $w$ that come as
the first and last words of the text. 
}

\subsection{Results}

We now aim to compare the frequency $f(w)$ with the inverse spatial
period $g(w)$ in each half of a given text.  We focus on sufficiently
frequent words, because $g(w)$ is not well-defined for words that appear
only once. Hence for a given half of a text we define the normalized
distance between $f(w)$ and $g(w)$:
\BEA
\label{ort}
\mu = \frac{1}{N_\Omega}\sum_{w\in \Omega} |f(w)-g(w)|,
\EEA
where $N_\Omega$ is the number of elements in the set $\Omega$, and
where $\Omega$ includes words that appear $k$ times. Tables \ref{mu_156}
and \ref{mu_350} exemplify the behavior of (\ref{ort}) for three
selected values of $k$: 15, 20 and 30. 
(We did not choose smaller values of $k$, since 
the definition of $1/g(w)$ as the average period becomes unclear.)
It is seen that the normalized
distance is typically larger in the first half,
\BEA
\mu_1>\mu_2,
\EEA
and that this effect gets stronger|both in terms of the percentage of
cases and the value of $W$ in (\ref{gnu2}), when the the set $\Omega$ in
(\ref{ort}) is restricted to common words of both halves; see Tables
\ref{mu_156} and \ref{mu_350}. 

\section{Features that do not show statistically significant
differences between two halves}
\label{nego}

So far we mostly concentrated on one level of textual hierarchy, i.e.
words.  Letters are on the hierarchy level below that of words. For the
total number of letters $L$ in each half the statistical evidence we got
is weaker, since the percentage of cases, where $L_1>L_2$ (i.e. the
first half has a larger overall number of letters than the second half)
and the W-statistics for $L_1>L_2$ are close to their critical values;
see Tables \ref{n_156} and \ref{n_350}. Moreover, in one of our datasets
the W-test is passed, while in the other it is not. However, there is a
weak, but a definite evidence for the validity of $L_1>L_2$. First, after
a random shuffling of texts, the percentage of cases where $L_1>L_2$ holds
drops down from its value $\simeq 0.58$ (for original texts) to $\simeq 0.5$
for shuffled texts; see Tables \ref{shuffle_156} and \ref{shuffle_350}. Second, 
$L_1>L_2$ shows significant correlations with (\ref{udosh}) and (\ref{cov});
see Tables \ref{corr_156} and \ref{corr_350}. Third, the relation $L_1>L_2$
holds in average for both datasets: 
\BEA
\label{guppi}
\frac{1}{156}\sum_{k=1}^{156} (L_1^{[k]}-L_2^{[k]})= 146.26,\qquad
\frac{1}{350}\sum_{k=1}^{350} (L_1^{[k]}-L_2^{[k]})= 340.97.
\EEA

\comment{Further distinction between words of the text can be made via
their average length (in letters): content words|which express specific
meaning|are normally longer than functional words that mostly serve for
establishing grammatic connections \cite{zipf}. }

For hierarchy levels higher than that of words, our results are
negative, i.e. they do not indicate on a statistically significant
difference between the halves. The number of sentences $\sigma$ does not
show significant differences; see Tables \ref{n_156} and \ref{n_350}.
Here we (conventionally) defined the sentence as the shortest sequence
of words located in between of any of the following symbols: comma, dot,
semicolon, question mark, exclamation mark. We also studied the number
of sentences, when the comma is excluded from the above list (not shown
in tables). This did not change our conclusion. 

We also calculated the full distribution of sentences over the
length (measured in words): $\kappa_\alpha$ the fraction of sentences with
word-length $\alpha$ ($\sum_\alpha\kappa_\alpha=1$). Two specific
characteristics of this distribution were looked at: the average
$\overline{\alpha}$, dispersion $\overline{\Delta(\alpha^2)}$ and
entropy $\ve$:
\begin{gather}
\label{deviation}
\overline{\alpha}={\sum}_\alpha\kappa_\alpha\alpha, \qquad
\overline{\Delta(\alpha^2)}
={\sum}_\alpha\kappa_\alpha
(\alpha-\overline{\alpha})^2.
\end{gather}
None of these quantities shown a statistically significant difference
between the halves; see Tables \ref{n_156} and \ref{n_350}.  Another
level of the textual hierarchy is the one containing paragraphs.
Denoting the number of paragraphs as $\rho$, we saw that there is no
statistical evidence in favor of $\rho_1>\rho_2$ or $\rho_1<\rho_2$; see
Tables \ref{n_156} and \ref{n_350}. Our results on Yule's constant that
describes the repetitiveness of words (see Appendix \ref{yu} details of
the definition) also do not indicate on a significant difference between
the halves. 

\section{Outlook}

We proposed a set of relations between statistical features of the two
halves of a meaningful text; see Table~\ref{tab0} for a summary of our
results. The validity of these relations is statistical, i.e. the
majority of them holds with $5\sigma$ significance of the Wilcoxon test;
see Appendix \ref{tablo}. No understanding of the text (or even knowing
its writing system) is needed for checking these relations. We
explicitly confirmed that all these relations disappear after a random
permutation of words in the text. 

We conjecture that these relations between the halves are connected to a
specific, information-carrying structure of the text, where the
information is introduced (defined) in the first half, and then is processed
in the second half. Such a structure is anticipated in text linguistics,
where the flow of the text narrative is conventionally separated between
the exposition and the denouement, which are typically located in the
first and second halves, respectively \cite{hutchins,valgina,hasan}.
This is however a qualitative concept, and hence the connection between
our results and the exposition-denouement is stated as a
hypothesis. Work is currently in progress for designing specific tests
for checking the hypothesis. 

Practically, knowing whether a string of symbols is a meaningful text or
not can be useful in cryptography, fraud detection and historical
analysis. The latter can refer to inferring whether a given text in
unknown writing system is meaningful or asemic \cite{writing_systems}.
One interesting application relates to the CETI problem (communication
with extra-terrestial intelligence) \cite{ceti}. Here the code of a
potential signal is completely unknown, but it can be plausibly
conjectured that a meaningful signal has similar differences between the
halves. (Fractioning into ``words'' is is possible, once the ``space
symbol'' is identified as the most frequent symbol in the text
\cite{ceti}.)

Some of our results were sporadically observed in literatures.
Ref.~\cite{minn} emphasized the translation invariance of books, but
still noted on a concrete text that its last part has less rare words.
Ref.~\cite{lrc_manin} noted the following (non-topical) differences
between the first and the second halves of {\it Moby Dick} (by H.
Melville): {\it (1.)} The word {\it is} is more frequent than {\it was}
in the first half, but less frequent in the second half. {\it (2.)} The
ratio of articles {\it the} to {\it a} is larger in the second half,
which may mean that the second half makes more concrete statements.

\comment{for PRE version:
After the initial version of this paper, one of its unknown referees
informed us that he/she counted the number of rare words in the
first/second halves of {\it all} of Project Gutenberg books that have
between 20000 and 40000 words in total (within the lengths of the texts
chosen by us). For a total of 10364 books, 5649 (54 \%) had more rare
words in the first half compared to the second half. We note that this
result has a statistical validity higher than $5\sigma$, since $0.5 +
5/(2 \sqrt{10364})\approx 0.5246$. For our situation the corresponding
percentages of rare words differences are larger by some 10-15 \%, see
Tables \ref{rare_156} and \ref{rare_350}, because we excluded from
consideration all those books that are {\it a priori} meaningless to
divide into two halves (hence such texts contribute into the noise):
biographies, poems, collections of short stories or essays (in
particular, folk stories), lectures, proceedings, letters. }

Our results concerning the compressibility features|in particular, the
result in section \ref{invo} on the compressibility decrease under
inverting the word order|are especially worth studying in more detail.
While we focused on the Lempel-Ziv complexity and the related zip method
for defining the compressibility, it is known that the Lempel-Ziv complexity
for long but finite sequences has a drawback of not capturing the real
randomness, i.e. not agreeing with the Kolmogorov complexity
\cite{cuba}. Hence more refined compression methods are to be studied in
future, e.g. the Huffman coding that is algorithmically slower, but does
capture the notion of randomness. It is also important to clarify whether
the compressibility difference between the original and inverted text 
can serve for quantifying syntagmatic correlations between the words; 
cf.~Footnote \ref{syntagma}.

Another important open problem relates to modeling the above effects. A
superficial modeling would be possible via altering the existing
sequential text-generating models (see \cite{simon,zane,kanter} for
example) such that e.g. they generate less rare words towards the end of
the text. But we should warn the reader against such quick attempts.
First, the main drawback of sequential models is that they do not
describe sufficiently well the distribution of rare words \cite{minn},
which was so far possible only via non-sequential statistical models
\cite{pre}. So a good model should predicts {\it both} the distribution
of rare words and their difference between the halves. Second, the
example of the Zipf law|with its numerous models and explanations
\cite{shrejder,li,simon,zane,kanter,hill,pre,liu,baek,vakarin,dover,latham,mandelbrot,mandel,manin}|shows
that excessive modeling even with a reasonable quantitative agreement is
not at all a guarantee for understanding the actual meaning of a complex
textual phenomenon. Hence at the present stage of our understanding we
want to concentrate on conceptual issues (and novel tests) relating
statistics to meaningfulness, more than to develop a purely statistical
model for explaining above results. 

\section*{Acknowledgements}

W. Deng was partially supported by the Fundamental Research Funds for
the Central Universities, the Program of Introducing Talents of
Discipline to Universities under grant no. B08033, and National Natural
Science Foundation of China (Grant Nos. 11505071 and 11905163). A.E. Allahverdyan was
supported by SCS of Armenia, grants No. 18RF-015 and No. 18T-1C090.

\appendix

\section{Tables}
\label{tablo}

\begin{table}[!htbp]
	\centering
	\caption{
Results for the first dataset of 156 texts. Notations:
$n$ is the number of different words; $n_1$ and $n_2$ refer to the first and second half, respectively. 
$c_1$ ($c_2$) is the overall frequency of common words in the first (second) half, 
$L$ is the number of
letters, $\rho$ is the number of paragraphs, $K$ is Yule's constant. Eqs.~(\ref{deviation}) define 
$\overline{\alpha}$ and $\Delta \alpha$. The number of sentences is denoted by $\sigma$.
\\ This table shows the W-value (\ref{gnu2}) for the relation $n_1>n_2$ (and other
indicated relations), and also the percentage with which this relation holds in the dataset. The W-test is passed 
according to the $3\sigma$ rule if $W>3\sigma_W(156)\approx 3391.02$; see (\ref{sigma}). 
The $3\sigma$ threshold for the percentage is $0.5+\frac{3}{2\sqrt{156}}\approx 0.62$. Table shows whether the $p\,\sigma$ 
rule is passed with $3\leq p\leq 5$. False means that even $3\sigma$ rule is not passed. 
\comment{ $3\sigma_W(156)=3391.02$, $4\sigma_W(156)=4521.36$, $5\sigma_W(156)=5651.7$
          $3\sigma_W(350)=11365.6$, $4\sigma_W(350)=15154.1, $5\sigma_W(350)=18942.7$
}
}
	\begin{tabular}{|l|c|c|c|}
		\hline\hline
		 & $W$ & $|W|>p\,\sigma_W$ & $\%$ \\
		 \hline
		$n_1>n_2$ & 6978 & $p=5$ & 0.724359 \\
   		$c_1<c_2$ & 5268 & $p=4$ & 0.705 \\
        \hline
		$L_1>L_2$ & 2292 & False & 0.589744 \\
       $\rho_1>\rho_2$ & -1312 & False& 0.403846 \\
                $K_1<K_2$ & -670    & False& 0.50641 \\
$\overline{\alpha}_1<\overline{\alpha}_2$ & 512  & False& 0.435897 \\
$\Delta\alpha_1<\Delta\alpha_2$ & 1708  & False& 0.410256 \\
 $\sigma_1>\sigma_2$ & -192  & False& 0.435897 \\
		\hline
	\end{tabular}
\label{n_156}
\end{table}

\begin{table}[!htbp]
	\centering
	\caption{Results for the second dataset of 350 texts. The same notations as in Table \ref{n_156}. The W-test is passed 
according to the $3\sigma$ rule if $W>3\sigma_W(350)\approx 11365.6$. The $3\sigma$ threshold for percentages is
	$0.5+\frac{3}{2\sqrt{350}}\approx 0.5802$. Table shows whether the $p\,\sigma$ 
rule is passed with $3\leq p\leq 5$. False means that even $3\sigma$ rule is not passed.}
	\begin{tabular}{|l|c|c|c|}
		\hline\hline
		 & $W$ & $|W|>p\,\sigma_W$ & $\%$ \\
		 \hline
		$n_1>n_2$ & 24030 & $p=5$ & 0.64 \\
   		$c_1<c_2$ & 13587 & $p=3$ & 0.583 \\
		$L_1>L_2$ & 15523 & $p=4$ & 0.594286 \\
        \hline
        $\rho_1>\rho_2$ & -5912 & False& 0.46 \\
        $K_1<K_2$ & -8467 & False& 0.571429 \\
		$\overline{\alpha}_1<\overline{\alpha}_2$
		                & 2891  & False& 0.5 \\
	$\Delta\alpha_1<\Delta\alpha_2$
	                        & 9007  & False& 0.428571 \\
  $\sigma_1>\sigma_2$ & -1008  & False& 0.5 \\
		\hline
	\end{tabular}
    \label{n_350}
\end{table}

\begin{table}[!htbp]
	\centering
        \caption{Random shuffle for the first dataset of 156 texts 10 shuffles; cf.~Table \ref{n_156}. 
                 For $h$ and $\frac{h}{n}$, the thresholds for appearance numbers of rare words are set to 3.
                 False means that even $3\sigma$ rule is not passed. }
	\begin{tabular}{|l|c|c|c|}
		\hline\hline
		& $W$ & $|W|>3\sigma_W$ & $\%$\\
		\hline
		$n_1>n_2$ & 213.   & False & 0.507692 \\
	    $L_1>L_2$ & -255.9 & False & 0.49359 \\
		$h_1>h_2$ & -42.2  & False & 0.489744 \\
        $\frac{h_1}{n_1}>\frac{h_2}{n_2}$ & 254.4  & False & 0.502564 \\ 
		$K_1<K_2$ & 690.4  & False & 0.485897  \\
		\hline
	\end{tabular}
\label{shuffle_156}
\end{table}

\begin{table}[!htbp]
	\centering
        \caption{Random shuffle for the second dataset of 350 texts averaged over 10 shuffles; cf.~Table \ref{n_350}. 
                 For $h(\kappa)$ and $\frac{h(\kappa)}{n}$, the thresholds for appearance numbers of rare words are 
                 set to $\kappa=3$; see (\ref{hh}--\ref{boris}).
                 False means that even $3\sigma$ rule is not passed.}
	\begin{tabular}{|l|c|c|c|}
		\hline\hline
		& $W$ & $|W|>3\sigma_W$ & $\%$\\
		\hline
		$n_1>n_2$ & -1978.2 & False & 0.484571 \\
	    $L_1>L_2$ & 112.2   & False & 0.501714 \\
		$h_1>h_2$ & 469.3   & False & 0.498857 \\
        $\frac{h_1}{n_1}>\frac{h_2}{n_2}$ & 1273.2  & False & 0.512    \\ 
		$K_1<K_2$ & 206     & False & 0.495143 \\
		\hline
	\end{tabular}
\label{shuffle_350}
\end{table}

\begin{table}[!htbp]
\centering
	\caption{Results for $h(\kappa)$ and $h(\kappa)/n$ for the first dataset of 156 texts; see (\ref{hh}--\ref{boris}). 
	The rareness is defined by $\kappa=1,...,5$.
	We demonstrate the percentage of cases, where the first half has a larger number of rare words. 
    The $3\sigma$ threshold for this percentage is $0.5+\frac{3}{2\sqrt{156}}\approx 0.62$.  
	We show the $W$ value (\ref{gnu2}) for the $W$-test. The test is 
    passed according to the $5\sigma$ rule for all but one case.}
\begin{tabular}{|l|c|c|c|c|c|c|}
	\hline\hline
	& \multicolumn{3}{|c|}{$h$} & \multicolumn{3}{|c}{$h/n$} \\
	\hline
        & $W$ & $|W|>p\,\sigma_W$ & $\% h_1>h_2$ & $W$ & $|W|>p\,\sigma_W$ & $\% h_1/n_1>h_2/n_2$\\
	\hline
	$\kappa= 5$ & 6811 & $p=5$ & 0.717949 & 5242 & $p=4$ & 0.647436 \\
	$\kappa= 4$ & 6961 & $p=5$ & 0.730769 & 5770 & $p=5$ & 0.679487 \\
	$\kappa= 3$ & 7016 & $p=5$ & 0.724359 & 6140 & $p=5$ & 0.711538 \\
	$\kappa= 2$ & 7095 & $p=5$ & 0.730769 & 6810 & $p=5$ & 0.717949 \\
	$\kappa= 1$ & 7246 & $p=5$ & 0.74359  & 6400 & $p=5$ & 0.705128 \\
	\hline
\end{tabular}
\label{rare_156}
\end{table}

\begin{table}[!htbp]
\centering
	\caption{Results for $h(\kappa)$ and $h(\kappa)/n$ for the second dataset of 350 texts. 
	The rareness is defined by $\kappa=1,...5$; see (\ref{hh}--\ref{boris}).
	We demonstrate the percentage of cases, where the first half has a larger 
    number of rare words. The $3\sigma$ threshold for percentages is
	$0.5+\frac{3}{2\sqrt{350}}\approx 0.5802$.  
	We show the $W$ value for the $W$-test. The test is 
    passed according to the $5\sigma$ rule for all but one case.} 
\begin{tabular}{|l|c|c|c|c|c|c|}
	\hline\hline
	& \multicolumn{3}{|c|}{$h$} & \multicolumn{3}{|c|}{$h/n$} \\
	\hline
        & $W$ & $|W|>p\,\sigma_W$ & $\% h_1>h_2$ & $W$ & $|W|>p\,\sigma_W$ & $\% h_1/n_1>h_2/n_2$\\
	\hline
	$\kappa=  5$ & 23657 & $p=5$ & 0.64     & 19207 & $p=5$ & 0.611429 \\
	$\kappa=  4$ & 23757 & $p=5$ & 0.64     & 20409 & $p=5$ & 0.631429 \\
	$\kappa=  3$ & 23455 & $p=5$ & 0.637143 & 19059 & $p=5$ & 0.614286 \\
	$\kappa=  2$ & 23393 & $p=5$ & 0.631429 & 18343 & $p=4$ & 0.58 \\
	$\kappa=  1$ & 23991 & $p=5$ & 0.651429 & 19285 & $p=5$ & 0.62 \\
	\hline
\end{tabular}
\label{rare_350}
\end{table}

\begin{table}[!htbp]
\centering
\caption{Shows the percentage and W-value for inequalities (\ref{urartu}).} 
\begin{tabular}{|l|c|c|c|c|c|c|}
	\hline\hline
	& \multicolumn{3}{|c|}{First dataset of 156 texts} & \multicolumn{3}{|c|}{Second dataset of 350 texts} \\
	\hline
    & $W$ & $|W|>p\,\sigma_W$ & \%        & $W$ & $|W|>p\,\sigma_W$ & \%  \\
	\hline
	  $C<n_1/2$  & 9805 & $p=5$ & 0.8397     & 33071 & $p=5$ & 0.7143 \\
	\hline\hline
	  $C<n_2/2$ & 7048 & $p=5$ & 0.7372      & 15625 & $p=4$ & 0.5943 \\
	\hline\hline
\end{tabular}
\label{urartu_t}
\end{table}

\begin{table}[!htbp]
\centering
\caption{The compressibility difference between the two halves; cf.~(\ref{sss}) 
and discussion around. It shows the percentage and W-value for the relation $s_1<s_2$. 
The relative compressibilities in (\ref{sss}) were calculates via the LZ-complexity (LZ) and the zip method.
The results are similar and confirm (\ref{sss}).
} 
\begin{tabular}{|l|c|c|c|c|c|c|}
	\hline\hline
	& \multicolumn{3}{|c|}{First dataset of 156 texts} & \multicolumn{3}{|c|}{Second dataset of 350 texts} \\
	\hline
           & $W$ & $|W|>p\,\sigma_W$ & $\% s_1<s_2$ & $W$ & $|W|>p\,\sigma_W$ & $\% s_1<s_2$\\
	\hline
	LZ     & 6962  & $p=5$ & 0.7051     & 17345 & $p=4$ & 0.5943 \\
	\hline\hline
    zip    & 6446 & $p=5$ & 0.7050      & 18539 & $p=4$ & 0.5940 \\
	\hline\hline
\end{tabular}
\label{zip}
\end{table}

\begin{table}[!htbp]
\centering
\caption{Shows the percentage and W-value for the relation (\ref{gel}), i.e. for the compressibility difference between a
text and its inverted version. Relation (\ref{gel}) is checked via the Lempel-Ziv (LZ) complexity and via the zip method.
} 
\begin{tabular}{|l|c|c|c|c|c|c|}
	\hline\hline
	& \multicolumn{3}{|c|}{First dataset of 156 texts} & \multicolumn{3}{|c|}{Second dataset of 350 texts} \\
	\hline
        & $W$ & $|W|>p\,\sigma_W$ & \% (\ref{gel})  & $W$ & $|W|>p\,\sigma_W$ & \% $s>s_{\rm inverted}$ \\
	\hline
	  LZ  & 12224 & $p=5$ & 0.9808     & 61095 & $p=5$ & 0.9743 \\
	\hline\hline
	  zip & 11512 & $p=5$ & 0.8846    & 40917 & $p=5$ & 0.777 \\
	\hline\hline
\end{tabular}
\label{inversion}
\end{table}

\begin{table}[!htbp]
	\centering
	\caption{Results for $\mu$ for the first dataset of $156$ texts. Here n.a. 
is the minimal number of appearances for words included into the definition (\ref{ort}) of $\mu$, e.g. ${\rm n.a.}>15$
means that only words with frequency $\geq 15/N$ are included, where $N$ is the total number of words in the text. }
\begin{tabular}{|l|c|c|r|c|c|r|}
	\hline\hline
	$\mu$ & \multicolumn{3}{|c|}{Words} & \multicolumn{3}{|c}{Common words} \\
	\hline
	n.a. & $W$ & $|W|>p\,\sigma_W$ & $\%\mu_1<\mu_2$ & $W$ & $|W|>p\,\sigma_W$ & $\%\mu_1>\mu_2$ \\
	\hline
	$>15$ & 5216  & $p=4$ & 0.685897 & 6340 & $p=5$ & 0.685897 \\
	$>20$ & 4774  & $p=4$ & 0.666667 & 5770 & $p=5$ & 0.692308 \\
	$>30$ & 5462  & $p=4$ & 0.679487 & 6066 & $p=5$ & 0.692308 \\
	\hline\hline
\end{tabular}
\label{mu_156}
\end{table}

\begin{table}[!htbp]
	\centering
	\caption{Results for $\mu$ for the second dataset of 350 texts; cf.~Table \ref{mu_156}.}
\begin{tabular}{|l|c|c|r|c|c|r|}
	\hline\hline
	$\mu$ & \multicolumn{3}{|c|}{Words} & \multicolumn{3}{|c}{Common words} \\
	\hline
	n.a. & $W$ & $|W|>3\sigma_W$ & $\%\mu_1<\mu_2$ & $W$ & $|W|>3\sigma_W$ & $\%\mu_1>\mu_2$ \\
	\hline
	$>15$ & 18443 & $p=4$ & 0.631429 & 30033 & $p=5$ & 0.7      \\
	$>20$ & 22401 & $p=5$ & 0.637143 & 32701 & $p=5$ & 0.722857 \\
	$>30$ & 25179 & $p=5$ & 0.665714 & 31053 & $p=5$ & 0.705714 \\
	\hline\hline
\end{tabular}
\label{mu_350}
\end{table}

\begin{table}[!htbp]
	\centering
	\caption{Pearson correlation coefficients between the number of appearances $\nu$
for various relations between the two halves in the second dataset of 156 texts; see also Appendix \ref{pearson}
in this context. Now $\nu([h_1>h_2])$ is the number of times (among 156 texts) that the relation $h_1>h_2$ holds; cf.~(\ref{cov}).
For $h$ we understand the number of words that appear once, while for $\mu$ the number of appearances of words included 
in (\ref{ort}) is set to $\geq 15$. Below $\mu_{\mathrm{C},1}$ and $\mu_{\mathrm{C},2}$ denote 
(\ref{ort}) applied to common words of (resp.) first and second halves. Significant correlations are underlined (i.e. the
coefficient is larger than $0.4$). For the relation $s_1<s_2$ [cf.~(\ref{sss})] we employed the zip compression method.}
	\begin{tabular}{|c|cccccc|}
		\hline\hline
		 & $\nu({[h_1>h_2]})$ & $\nu({[L_1>L_2]})$ & $\nu({\left[\frac{h_1}{n_1}>\frac{h_2}{n_2}\right]})$ 
                                & $\nu({[\mu_1>\mu_2]})$ & $\nu({[\mu_{\rm C,1}>\mu_{\rm C, 2}]})$ & $\nu({[s_1<s_2]})$ \\
		\hline
  $\nu({[n_1>n_2]})$  & \underline{0.820515} & \underline{0.477115} & \underline{0.513464} & -0.200699 &-0.138885 & \underline{0.482} \\
  $\nu({[h_1>h_2]})$  & -- & \underline{0.435441} & \underline{0.682702} & -0.171928 & -0.108672 & \underline{0.457} \\
  $\nu({[L_1>L_2]})$  & -- & -- & 0.260886 & -0.0813513 & -0.0532743 & 0.061 \\
  $\nu({\left[\frac{h_1}{n_1}>\frac{h_2}{n_2}\right]})$ & -- & -- & -- & -0.168133 & -0.107558 & 0.383 \\
  $\nu({[\mu_1>\mu_2]})$ & -- & -- & -- & -- & \underline{0.791722} & 0.04698 \\
 \hline\hline
	\end{tabular}
\label{corr_156}
\end{table}

\begin{table}[!htbp]
	\centering
	\caption{Pearson correlation coefficients between the number of appearances $\nu$
for various relations between the two halves in the second dataset of 350 texts. For notations see Table \ref{corr_156}.}
	\begin{tabular}{|c|cccccc|}
		\hline\hline
		 & $\nu({[h_1>h_2]})$ & $\nu({[L_1>L_2]})$ & $\nu({\left[\frac{h_1}{n_1}>\frac{h_2}{n_2}\right]})$ 
                                               & $\nu({[\mu_1>\mu_2]})$ & $\nu({[\mu_{\rm C,1}>\mu_{\rm C, 2}]})$ & $\nu({[s_1<s_2]})$\\
		\hline
  $\nu({[n_1>n_2]})$  & \underline{0.850415} & \underline{0.495557} & \underline{0.504262} & -0.105619 & -0.0805328 & \underline{0.568} \\
  $\nu({[h_1>h_2]})$  & -- & \underline{0.433519} & \underline{0.650251} & -0.112285 & -0.109906 & \underline{0.519} \\
  $\nu({[L_1>L_2]})$  & -- & -- & 0.204271 & -0.0683025 & -0.0812633 & {0.182} \\
  $\nu({\left[\frac{h_1}{n_1}>\frac{h_2}{n_2}\right]})$ & -- & -- & -- & -0.109572 & -0.104045 & \underline{0.42}\\
  $\nu({[\mu_1>\mu_2]})$ & -- & -- & -- & -- & \underline{0.611309} & 0.0803 \\
 \hline\hline
	\end{tabular}
\label{corr_350}
\end{table}

%\clearpage

\section{Number of words with a fixed frequency and Yule's constant}
\label{yu}

Let us recall how Yule's constant is defined \cite{baa}.
Define $V_m$ to be the number of words that (in a fixed text) appear $m$
times. We get two obvious features:
\begin{equation}
\label{kush}
{\sum}_{m=1}^{f_1 N}\, V_m=n, ~~{\sum}_{m=1}^{f_1N} \, mV_m=N,
\end{equation}
where $n$ and $N$ are, respectively, the number of different words and
the number of all words, and $f_1N$ is the number of times the most
frequent word appears in the text.  Note that for a sufficiently small
$m$, $V_m$ is either zero or one. For instance, consider the text {\it
The Age of Reason} by T. Paine, 1794 (the major source of British
deism), whose first half has $N=11612$ total words and $n=2012$ different
words.  In the first half of this text: $V_{f_1N}=V_{929}=1$,
$V_{928\geq m\geq 575}=0$, $V_{574}=1$, $V_{573\geq m\geq 388}=0$,
$V_{387}=1$ {\it etc}. 

Take a word $w$ that appears $m$ times in a text with length $N$. Now
$\frac{m}{N}$ is the probability that a randomly taken word in the text
will be $w$. Likewise, $\frac{m(m-1)}{N(N-1)}$ is the probability that
the second randomly taken word in the text will be again $w$. Both probabilities
refer to a word $w$ that appears $m$ times. The probability to take
such a word among $n$ distinct words of the text is $\frac{V_m}{n}$; cf.~(\ref{kush})
\footnote{Hence one can define the entropy $-\sum_m\frac{V_m}{n}\ln \frac{V_m}{n}$ that
characterizes the inhomogeneity of distribution of distinct words.  
This entropy was employed in \cite{cohen_fractal} for distinguishing between
natural and artificial texts.}.
Thus the average $\frac{1}{n}{\sum}_{m=1}^{f_1 N}\left[
V_m\frac{m(m-1)}{N(N-1)}\right]$ is a measure of repetitiveness of
words. The Yule's constant $K$ employs this quantity without the factor
$\frac{1}{n}$, since it wants to have something weakly dependent on $N$
\cite{baa}.  For us this feature is not important, since we compare the
halves of a text.  Following the tradition, we also omit the factor
$\frac{1}{n}$, but we stress that including it does not change our
conclusions. Using (\ref{kush}) and $N\simeq N-1\gg 1$, the Yule's
constant reads
\cite{baa}
\begin{equation}
\label{yule}
K=10^2\left[-\frac{1}{N}+
{\sum}_{m=1}^{f_1 N}\, V_m\,\frac{m^2}{N^2}\right],
\end{equation}
where $10^2$ is a conventional factor we applied to keep $K={\cal
O}(1)$.

\section{Pearson's correlation coefficient}
\label{pearson}

We have $M$ texts ($k=1,...,M$ for our case $M=350$ or $M=156$) and 
5 features ($a=1,...,5$):

1. $n_1>n_2 $

2. $L_1>L_2$

3. $h_1>h_2$

4. $h_1/n_1>h_2/n_2$

5. space frequency.

Let us define $\nu_k^{[a]}$ such that $\nu_k^{[a]}=1$ ($\nu_k^{[a]}=0$) if the
feature $a$ holds (does not hold) for text $k$. Let us now define the
Pearson correlation coefficient between the features:
\BEA
\label{tsul}
\frac{\frac{1}{M} \sum_{k=1}^M  (\nu_k^{[a]} - \bar{\nu}^{[a]}) (\nu_k^{[b]} - \bar{\nu}^{[b]}) }
{\sqrt{\frac{1}{M} \sum_{k=1}^M  (\nu_k^{[a]} - \bar{\nu}^{[a]})^2}
\sqrt{\frac{1}{M} \sum_{k=1}^M  (\nu_k^{[b]} - \bar{\nu}^{[b]})^2}  },
\EEA
where 
\BEA
\bar{\nu}^{[a]}\equiv \frac{1}{M} \sum_{k=1}^M  \nu_k^{[a]}.
\EEA
Eq.~(\ref{tsul}) can be simplified as 
\BEA
\label{tsul2}
\frac{\frac{1}{M} \sum_{k=1}^M  \nu_k^{[a]} \nu_k^{[b]} -\bar{\nu}^{[a]} \bar{\nu}^{[b]} }
{\sqrt{\bar{\nu}^{[a]}(1-\bar{\nu}^{[a]})}
\sqrt{\bar{\nu}^{[b]}(1-\bar{\nu}^{[b]})}  },
\EEA

%\clearpage

\end{document}